\documentclass{article}

\usepackage[final]{nips_2017}

\usepackage[utf8]{inputenc} 
\usepackage[T1]{fontenc}    
\usepackage{hyperref}       
\usepackage{url}            
\usepackage{booktabs}       
\usepackage{amsfonts}       
\usepackage{nicefrac}       
\usepackage{microtype}      

\usepackage{subcaption} 
\usepackage{graphicx}

\usepackage[english]{babel} 

\usepackage{caption} 

\usepackage{times}
\usepackage{graphicx} 

\usepackage{sidecap} 
\sidecaptionvpos{figure}{c}

\usepackage{tikz}
\usetikzlibrary{positioning, shapes}
\usetikzlibrary{shapes.misc, fit}

\usepackage{natbib}
\usepackage{amssymb}
\usepackage{algorithm}
\usepackage{algorithmic}
\usepackage{amsmath}
\usepackage{mathtools}

\usepackage{hyperref}

\usepackage{multicol}
\usepackage{lipsum}
\usepackage{mwe}

\newcommand{\commentedbox}[2]{%
  \mbox{
    \begin{tabular}[t]{@{}c@{}}
    $\boxed{\displaystyle#1}$\\
    #2
    \end{tabular}%
  }%
}

\definecolor{color1}{RGB}{0.0, 50, 0.0} 
\definecolor{color2}{RGB}{0,20,20} 

\usepackage{hyperref} 
\hypersetup{hidelinks,colorlinks,breaklinks=true,urlcolor=color2,citecolor=color1,linkcolor=color1,bookmarksopen=false,pdftitle={Title},pdfauthor={Author}}

\title{Doubly Stochastic Adversarial Autoencoder}

\author{
  Mahdi~Azarafrooz \\
  Department of Research and Intelligence\\
  Cylance, Irvine, CA \\
  \texttt{mazarafrooz@cylance.com} \\
}

\begin{document}

\maketitle
Any autoencoder network can be turned into a generative model by imposing an arbitrary prior distribution on its hidden code vector. Variational Autoencoder (VAE) [2] uses a KL divergence penalty to impose the prior, whereas Adversarial Autoencoder (AAE) [1] uses {\it generative adversarial networks} GAN [3]. GAN trades the complexities of {\it sampling} algorithms with the complexities of {\it searching} Nash equilibrium in minimax games. Such minimax architectures get trained with the help of data examples and gradients flowing through a generator and an adversary.  
A straightforward modification of AAE is to replace the adversary with the maximum mean discrepancy (MMD) test [4-5]. This replacement leads to a new type of probabilistic autoencoder, which is also discussed in our paper.
We propose a novel probabilistic autoencoder in which the adversary of AAE is replaced with a space of {\it stochastic} functions. 
This replacement introduces a new source of randomness, which can be considered as a continuous control for encouraging {\it explorations}. This prevents the adversary from fitting too closely to the generator and therefore leads to more diverse set of generated samples. 

\section{Doubly Stochastic AAE}
We consistently refer to any random variable and their realization with calligraphic fonts and small letters respectively. 
Fix an autoencoder (AE) with encoder parameters $\theta \in \Theta$ that gets i.i.d data samples $\mathcal{X},\mathcal{X}_1,...,\mathcal{X}_N$ as input and outputs the corresponding latent vectors $\mathcal{Z},\mathcal{Z}_1,...,\mathcal{Z}_N$.  Assume we don't know the distribution of $\mathcal{X}$, but we want to impose an arbitrary distribution $\mathcal{P}$ on $\mathcal{Z}$.
One can impose $\mathcal{P}$ on $\mathcal{Z}$ through regularization of a standard AE by minimizing {\it any suitable discrepancy} between aggregated posterior $Q_{\theta}(\mathcal{Z})=\mathbb{E}_{\mathcal{X}} [Q_{\theta}(\mathcal{Z}|\mathcal{X})]
$ and any arbitrary prior $\mathcal{P}$. This can be formalized as the following optimization problem:
\begin{eqnarray}\label{Discrepency_optimization}
\begin{array}{l}
\begin{split} 
\underset{\theta  \in \Theta}{\mathrm{\min}} ~ \delta(\mathcal{P},Q_{\theta} (\mathcal{Z}))
\end{split}
\end{array}
\end{eqnarray}
where $\delta$ is a suitable discrepancy measure. For example, we end up with a maximum mean discrepancy AE (MMD-AE) using the following discrepancy measure:
\begin{eqnarray}\label{mmd-discrepency}
\begin{array}{l}
\begin{split} 
\delta_{MMD}(\mathcal{P},Q_\theta(\mathcal{Z}))=
\underset{f \in \mathcal{H}}{\mathrm{sup}} ~ \mathbb{E}[f(\mathcal{Y})] -\mathbb{E}[f(\tilde{\mathcal{Y}})]
\end{split}
\end{array}
\end{eqnarray}

where $f$ is a function living in a reproducing kernel Hilbert space RKHS $H$ and $\mathcal{Y} \sim \mathcal{P}$ and $\tilde{\mathcal{Y}} \sim Q_{\theta}(\mathcal{Z})$. The authors of [4-5] use a closed form expression for Eq.\ref{mmd-discrepency}.
This makes the implementation straightforward, but it hides the {\it minimax} nature of the problem.
If we replace Eq.\ref{mmd-discrepency}  in Eq.\ref{Discrepency_optimization} we are back to a minimax problem similar to that of adversarial networks as the following:
\begin{eqnarray}\label{mmd-minimax}
\begin{array}{l}
\begin{split} 
\underset{\theta}{\mathrm{\min}} ~ \underset{f \in \mathcal{H}}{\mathrm{sup}} ~  \commentedbox{\mathbb{E}_{\mathcal{Y} \sim \mathcal{P}}[f(\mathcal{Y})] -\mathbb{E}_{\tilde{\mathcal{Y}} \sim Q_{\theta}(\mathcal{Z})}[f(\tilde{\mathcal{Y}})]}{ \tiny $\delta(\mathcal{P},Q_{\theta}(\mathcal{Z}))$ }
\end{split}
\end{array}
\end{eqnarray}
Due to the reproducing property of $H$, the expectation of any function $f$ in RKHS $H$ with respect to random variable $\mathcal{Y}$  can be computed as an inner product with its so called  {\it kernel mean embedding $\mathbb{E}[k(\mathcal{Y},.)]$}:
\begin{eqnarray}\label{reproducing-property}
\begin{array}{l}
\begin{split} 
\mathbb{E}[f(\mathcal{Y})]=\langle f, \mathbb{E}[k(\mathcal{Y},.)] \rangle
\end{split}
\end{array}
\end{eqnarray}
where $k$ is the kernel associated with RKHS $H$.
Using Eq.\ref{reproducing-property} and Eq.\ref{mmd-minimax}, we note that the adversary's best response training dynamic is determined by the stochastic gradient terms $\frac{\partial \delta}{\partial f}=\xi(.) \coloneqq \underset{\mathcal{Y} \sim \mathcal{P}}{\mathbb{E}} [k(\mathcal{Y},.)]-\underset{\tilde{\mathcal{Y}} \sim Q_{\theta}(\mathcal{Z})}{\mathbb{E}} [k(\tilde{\mathcal{Y}},.)]$.
Since prior $\mathcal{P}$ is smooth enough, term $\underset{\tilde{\mathcal{Y}} \sim Q_{\theta}(\mathcal{Z})}{\mathbb{E}} [k(\tilde{\mathcal{Y}},.)]$ determines the smoothness of the stochastic gradients. Not smooth enough $Q_{\theta}(\mathcal{Z})$ implies non-smooth gradients and therefore a more difficult optimization problem. It can lead to degenerate solutions where the generator collapses into sampling only a few modes. This is especially important considering the deterministic functionality of the encoder. It is because the only source of randomness originates from the training data itself which may not be enough for smoothing out the aggregated posterior $Q_{\theta}(\mathcal{Z})$. We propose to massage the stochastic gradients $\xi(.)$ with the {\it extra} source of stochasticity $\mathcal{W}$. This leads to {\it doubly stochastic} gradient terms $\zeta(.) \coloneqq \mathbb{E}_{\mathcal{W}}\xi(.)$. 
The introduction of a new source of randomness to $\xi(.)$ is straightforward thanks to the existing duality between the Kernel and Random processes, as explained in the following Theorem:


\textbf{Theorem 1 [9]} \textit{Duality between Kernels and Random Processes}:  If $k(x,x')$ is a positive definite kernel, then there exits a set $\Omega$, a measure $\mathbb{P}$ on $\Omega$, and random function $\phi_{\mathcal{W}}(x):\mathcal{X}\rightarrow \mathbb{R}$ from $L_2(\Omega,\mathbb{P})$, such that $k(x,x')=\int_{\Omega} \phi_{\mathcal{W}}(x)\phi_{\mathcal{W}}(x')d\mathbb{P}(\mathcal{W})$.



\textbf{Example 1} For Gaussian RBF kernel, $k(x-x')=\exp(\|x-x'\|^2/2\sigma^2)$, $\phi_{\mathcal{W}}(x)=\exp(-i\mathcal{W}^{\top}x)$ and $\mathbb{P}(w)=\frac{\exp(-\|w\|_2^2/2)}{(2\pi)^{d/2}}$

As the result of Theorem 1, we can rewrite the massaged gradients $\zeta=[\phi_{\mathcal{W}}(\mathcal{Y})-\phi_{\mathcal{W}}(\tilde{\mathcal{Y}})] \phi_{\mathcal{W}}(.)$ with $\mathcal{W} \sim \mathbb{P}(\mathcal{W})$. 
Using a similar approach to Doubly Stochastic Kernel Machines [6], we can now approximate a new adversary with parameter $\alpha$ by the linear combination of doubly stochastic gradient terms $\zeta$ i.e. $f_{\alpha}(.)=\alpha\zeta(.)$.  The adversary affects the dynamic of training through the adjustment in $\alpha$ as is visualized in Fig.1. We refer to this deep generative architecture as doubly stochastic adversarial autoencoder (DS-AAE).

Beside the residual error of approximating adversary $f_{\alpha}$ with its first order gradient terms, there is another concern that needs to be addressed for DS-AAE. The key difference between $\zeta(.)$ and $\xi(.)$ is that $\zeta(.)$ could fall outside of the RKHS, but $\xi(.)$ is always in RKHS. This is due to the term $\phi_w(.)$ not being in RKHS. Fortunately, the proof of convergence to RKHS $H$ with a small enough learning rate is discussed in [6]. 
\begin{SCfigure}
  \centering
  \includegraphics[scale=0.25]{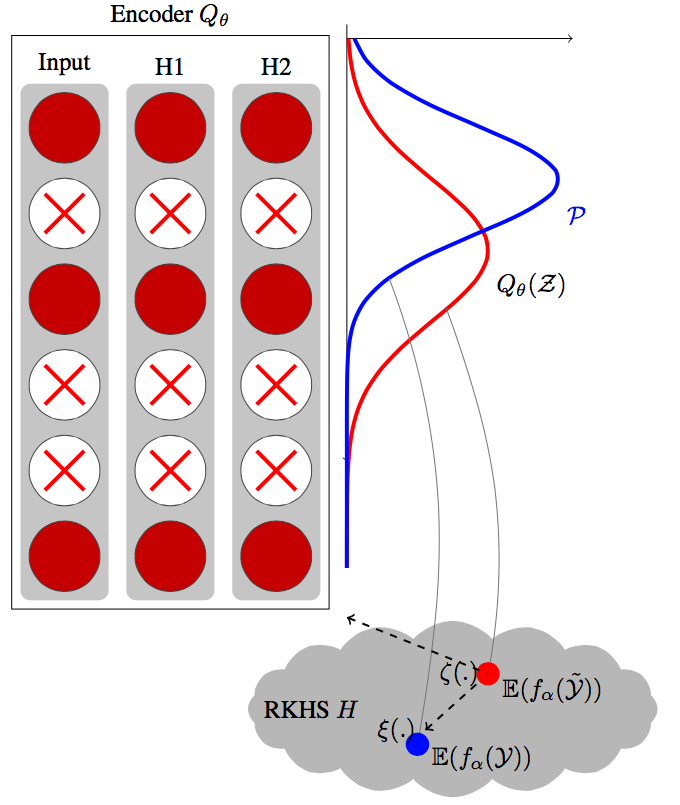}  
  \caption{\small \textbf{Scheme of DS-AAE} Each distribution is mapped into a reproducing kernel Hilbert space via an expectation operation. The generator strategy is to adjust the parameter of encoder $\theta$ to decrease the distance between the blue and red dot while the adversary's is to adjust $\alpha$ to increase the distance. Not enough stochasticity in $\mathcal{X}$ limits the adversary's power as is reflected in its stochastic gradient terms $\xi=\underset{\mathcal{Y} \sim Q_{\theta}(\mathcal{Z})}{\mathbb{E}} [k(\mathcal{Y},.)]$. With the introduction of a new source of randomness, the adversary gains extra power, as is reflected in its doubly stochastic gradient terms $\zeta=\mathbb{E}_{\mathcal{W}}\xi(.)$. This extra boost to the adversary, however, can lead to the search outside of $H$, where the gradients are no longer valid closed form expressions. Fortunately, [6] shows that with the small learning rates, the search returns back to $H$ and convergence is guaranteed.}
\end{SCfigure}


\section{Experiments}
We generate $\mathcal{W}$ according to Example 1 with a fixed seed and $\sigma=1$ for DS-AAE. The batch size is 1000. Our experiments show that DS-AAE is more sensitive to the batch size than the choice of $\sigma$. In the case of MMD-AE, a mixture of $\sigma$ values 2, 5, 10, 20, 40, 80 is used, similar to that of [5]. The encoder and decoder both have three layers of 1024, 512 and 216 hidden units with ReLU activation for every layer except the last layer of decoder that is a sigmoid activation function. Cross entropy is used for reconstruction loss. The prior $\mathcal{P}$ is Gaussian and the dimensionality of the hidden code is 6 for the DS-AAE and 4 for MD-AAE. The only used dropout is at the first input layer of the encoder with the rate of $20\%$. The initial learning rate for the reconstruction loss is adjusted at 0.001 and 0.001 for the adversarial architectures, followed by Adam stochastic optimization. Comparison of deep generative models performance is hard, especially for the log-likelihood free models [7]. Parzen window estimation of the log-likelihood is obtained by drawing 10K samples from the trained model on MNIST. The results are shown in Table 1. From the qualitative perspective, we can see from Fig. 2b, Fig. 2d, and the reported results in [1] that the drawn samples for both MMD-AE and AAE are more homogenous than DS-AAE.  In the case of DS-AAE, it is almost as if different persons were writing the digits in each panel. This quality test is also used in [8]. DS-AAE enjoys from extra randomness in the minimax optimization framework, which helps the generative model to explore multiple modes and mitigate the risk of collapse. The learned coding space of DS-AAE exhibits sharp transitions and has no ``holes'', as is shown in Fig.2a. This is similar to AAE and unlike VAE. However, it recovers more of a mixture of 2D-Gaussian distribution rather than a 2D-Gaussian. We leave further investigation of this interesting observation to a future version of this paper. 
\begin{figure}[t]
\centering
\begin{subfigure}{0.4\textwidth}
\centering
\includegraphics[width = \textwidth]{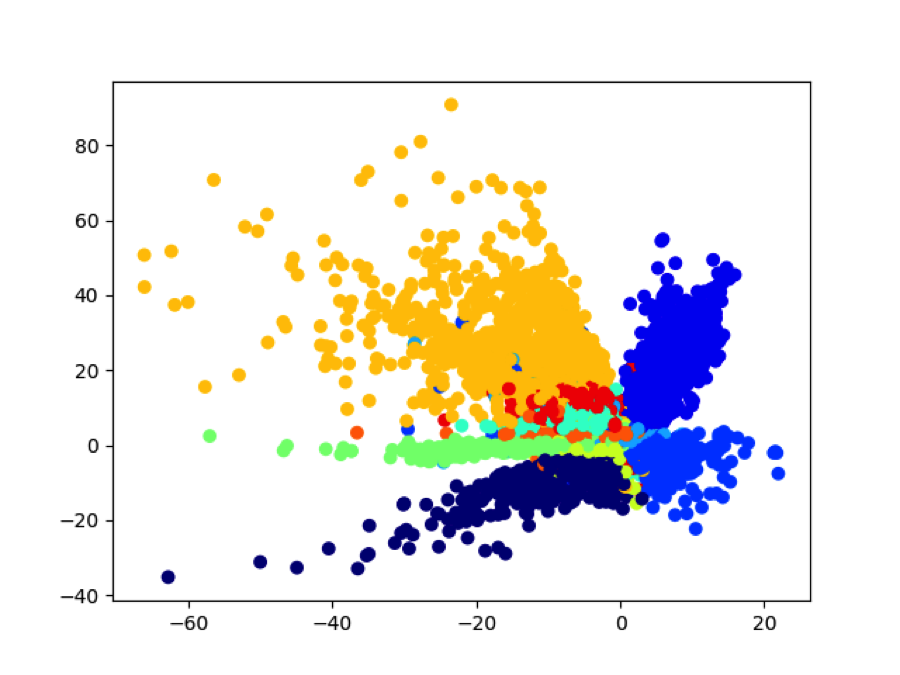}
\caption{\tiny DS-AAE: The hidden code $\mathcal{Z}$ of the hold-out images with two latent variables. Each color represents the associated label on MNIST data set. }
\label{fig:left}
\end{subfigure}
\begin{subfigure}{0.4\textwidth}
\centering
\includegraphics[width = \textwidth]{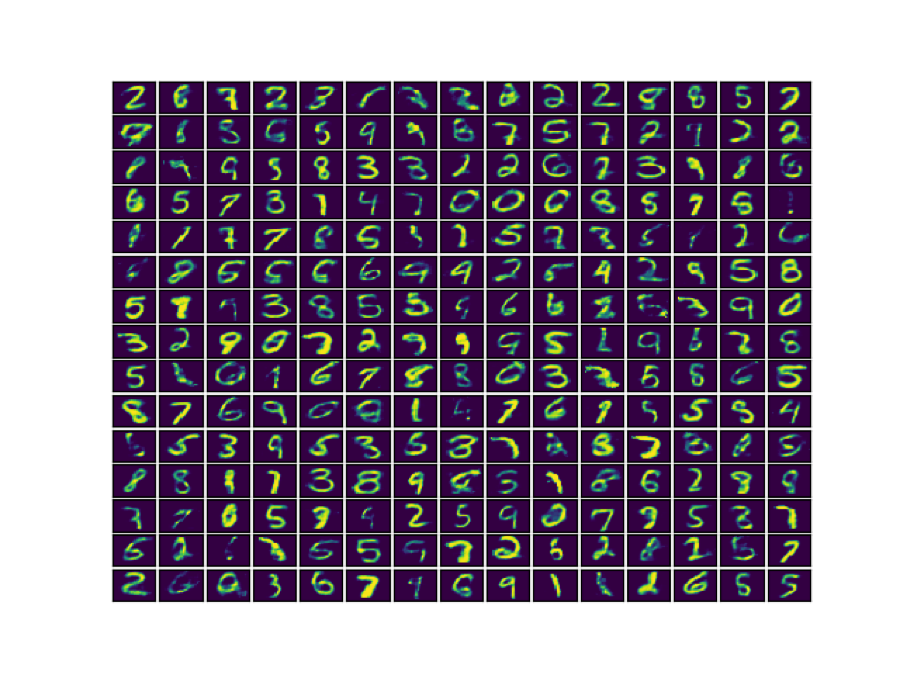}
\caption{\tiny DS-AAE: Drawn samples after 1000 epochs.}
\label{fig:right}
\end{subfigure}
\begin{subfigure}{0.4\textwidth}
\centering
\includegraphics[width = \textwidth]{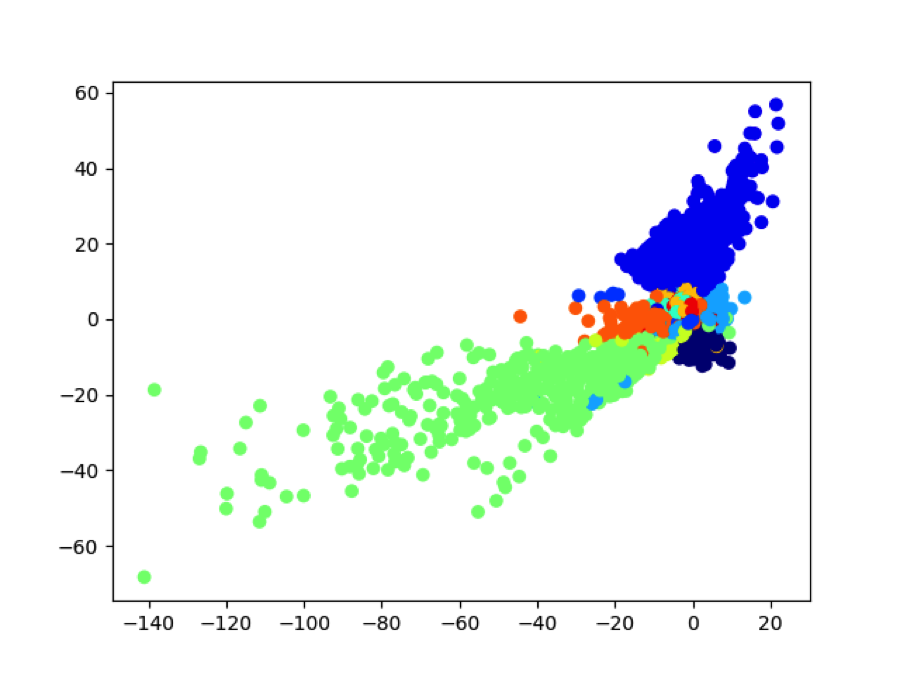}
\caption{\tiny MMD-AE: The hidden code $\mathcal{Z}$ of the hold-out images with 2 latent variables. Each color represents the associated label on MNIST data set. }
\label{fig:left}
\end{subfigure}
\begin{subfigure}{0.4\textwidth}
\centering
\includegraphics[width = \textwidth]{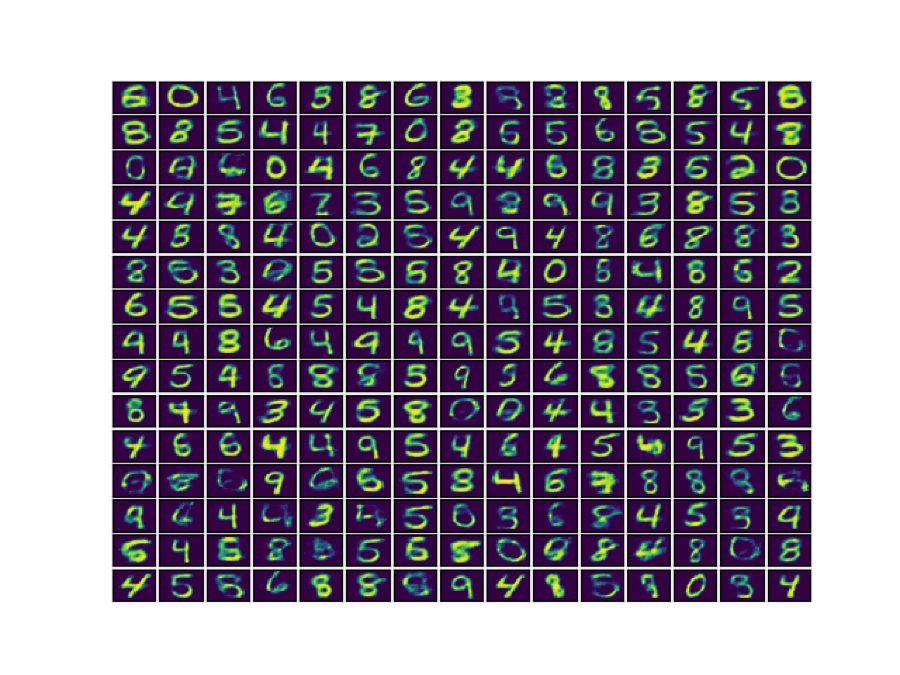}
\caption{\tiny MMD-AE: Drawn samples after 1000 epochs.}
\label{fig:left}
\end{subfigure}
\caption{\small Comparison between MMD-AE and DS-AAE on MNIST.}
\label{fig:combined}
\end{figure}

\begin{center}
\begin{table}[t]
\centering 
\begin{tabular}{ l  || c || c || c || c}
  \hline
  & MNIST \\
  \hline
  GAN [3] & $225 \pm 2$ \\
  GMMN + AE [5] & $282 \pm 2$ \\
  Adversarial Autoencoder [1] & $340 \pm 2$\\
 MMD-AE & $\mathbf{228 \pm 1.59}$  \\
  DS-AAE & $\mathbf{243.16 \pm 1.65}$ \\
  \hline
\end{tabular}
\caption{\label{table:table_mnist}\small Parzen window estimate of the log-likelihood obtained by drawing 10K samples from the trained model.}
\end{table}
\end{center}

\section{References}
\phantomsection
\bibliographystyle{unsrt}
\bibliography{sample}

[1] Alireza Makhzani and Jonathon Shlens and Navdeep Jaitly and Ian Goodfellow (2016) Adversarial autoencoders {\it International Conference on Learning Representations (ICLR)}

[2] Diederik P Kingma and Max Welling (2014). Auto-encoding variational bayes. {\it International Conference on Learning Representations (ICLR)}.

[3] Ian Goodfellow, Jean Pouget-Abadie, Mehdi Mirza, Bing Xu, David Warde-Farley, Sherjil Ozair, Aaron Courville, and Yoshua Bengio. (2014) Generative adversarial nets. {\it In Advances in Neural Information Processing Systems}, pp.\ 2672--2680.

[4] Gintare Karolina Dziugaite, Daniel M. Roy, and Zoubin Ghahramani (2015). Training generative neural networks via maximum mean discrepancy optimization. {\it In Proceedings of the Thirty-First Conference on Uncertainty in Artificial Intelligence (UAI'15), Marina Meila and Tom Heskes (Eds.). AUAI Press}, Arlington, Virginia, United States, pp.\ 258-267.

[5] Yujia Li, Kevin Swersky, and Richard Zemel. Generative moment matching networks (2015). {\it International Conference on Machine Learning (ICML)} . 

[6] Dai, Bo and Xie, Bo and He, Niao and Liang, Yingyu and Raj, Anant and Balcan, Maria-Florina F and Song, Le (2014). Scalable Kernel Methods via Doubly Stochastic Gradients.  {\it Advances in Neural Information Processing Systems 27}

[7] Lucas Theis, Aaron van den Oord, and Matthias Bethge (2015). A note on the evaluation of generative
models. {\it arXiv preprint arXiv:1511.01844}

[8] Saatchi, Y and Wilson, AG, (2017) Bayesian GAN.

[9] A. Devinatz. Integral representation of pd functions. (1953) {\it Trans. AMS 74(1)} pp. 56--77 

\end{document}